\documentclass{article}

\usepackage[final]{neurips_2022}

\usepackage{graphicx}

\usepackage[utf8]{inputenc} 
\usepackage[T1]{fontenc}    
\usepackage{hyperref}       
\usepackage{url}            
\usepackage{booktabs}       
\usepackage{amsfonts}       
\usepackage{nicefrac}       
\usepackage{microtype}      
\usepackage{xcolor}         
\usepackage{color}
\usepackage{wrapfig}
\usepackage{subfig}
\usepackage{graphicx}
\usepackage[ruled]{algorithm2e}

\title{Rank-One Editing of Encoder-Decoder Models}

\author{%
Vikas Raunak \quad Arul Menezes \\
Microsoft Azure AI\\
\texttt{\{viraunak,arulm\}@microsoft.com}\\}

\begin{document}

\maketitle

\begin{abstract}
Large sequence to sequence models for tasks such as Neural Machine Translation (NMT) are usually trained over hundreds of millions of samples. However, training is just the origin of a model's life-cycle. Real-world deployments of models require further behavioral adaptations as new requirements emerge or shortcomings become known. Typically, in the space of model behaviors, behavior deletion requests are addressed through model retrainings whereas model finetuning is done to address behavior addition requests, both procedures being instances of data-based model intervention. In this work, we present a preliminary study investigating rank-one editing as a direct intervention method for behavior deletion requests in encoder-decoder transformer models. We propose four editing tasks for NMT and show that the proposed editing algorithm achieves high efficacy, while requiring only a single instance of positive example to fix an erroneous (negative) model behavior.
\end{abstract}

\section{Introduction}

Large neural models with huge training costs run the risk of becoming software monoliths due to a lack of abstractions in debugging and editing such models. This presents a risk towards their utility in domains where rapid interactivity with the stakeholders is required, in the absence of which the models could lead to significant real-life costs or face barriers to adoption due to bespoke salient errors, despite having high average-case performance. However, post-training, direct interaction(s) with the model to change its behavior is relatively understudied for sequence to sequence tasks such as Neural Machine Translation (NMT). Simultaneously, a number of post-deployment problems faced by such models could be reduced to edit requests for fixing specific deleterious behaviors.

The task of model editing is closely related to the problems of machine unlearning \citep{machine_unlearning}, domain adaptation \citep{thompson-etal-2019-overcoming}, model patching \citep{patching} and continual learning \citep{continual}. However, while such problems focus on a specific model end state with respect to data or task performance (data deletion, performance improvement on a specific domain, performance improvement on a specific-task, incorporation of new knowledge, respectively), the focus of model editing is on correcting model behavior on specific sample \textit{instances} (not tasks or domains), while preserving as much of model's earlier behavior (general performance) as possible. Therefore, readily applying the techniques developed for those tasks to model editing requests is unsuitable owing to the new constraints imposed by the model editing problem. 

The first of these constraints is that the supplied data for the editing task comes in the form of a single instance that shows an incorrect model behavior, e.g., an input instance on which the model hallucinates or shows an error in the generated output. These behaviors naturally \textit{become} known post-training through  model use or behavioral testing as the trained model moves further in its life-cycle \citep{checklist, salted}. For arbitrary input-output instances depicting such negative behaviors, data-based interventions cannot be readily designed. Secondly, the editing operation is required to be computationally inexpensive, with the ratio of computation costs for model editing to model retraining required to remain extremely low for it to be effective in a viable manner. 


In this work, we investigate the recently proposed rank-one model editing algorithm \citep{bau_rewriting}, for encoder-decoder models on the canonical sequence to sequence task of NMT \citep{sutskever2014sequence, bahdanau2014neural, transformer}. While rank-one editing has been leveraged for editing classifiers \citep{santurkar_editing} as well as as language models \citep{bau_gpt3_editing}, we present the first study of its applicability in the case of encoder-decoder models. Specifically, we consider the task of NMT and propose a set of four editing tasks, upon which the model editing algorithm could be evaluated.  Our contributions are as follows:
\begin{enumerate}
    \item We show that localized model edits could be successfully applied for encoder-decoder models as well, with the \textit{efficacy} of edits closely tied to the specific \textit{location} of its application.
    \item We propose a set of four \textit{editing tasks} for NMT and demonstrate that \textit{directly} applying rank-one editing considerably degrades general model performance.
    \item We propose \textit{edit-dropout}, which randomly drops out edit update vectors, as a simple but effective technique to alleviate the drop in performance due to direct rank-one model editing.
\end{enumerate}

\section{Editing Tasks in Neural Machine Translation}
\label{editing_tasks}

We study model editing through the lens of four editing tasks, which include both the task of fixing \textit{isolated} model behaviors as well as tasks of fixing consistent \textit{patterns} of errors. The tasks are:

\begin{enumerate} 
    \item \textbf{Fixing Hallucinations}: In this task, an instance of a hallucinating sample (input-output pair) is presented and the the goal of the edit is to remove model hallucination on this input. Hallucinations represent an error mode which significantly reduces user trust in the models \citep{raunak-etal-2021-curious}. We collect the hallucinating instance (Figure \ref{fig:demo}b), for which the edit is applied, using the oscillatory hallucination detector described in \citet{raunak-etal-2021-curious}.   
    \item \textbf{Memorization Mitigation}: In this task, a memorized input is presented and the goal of the edit is to mitigate the memorization i.e. rectify the model to generate the non-memorized output. We collect the memorized instance using the \textit{extractive memorization} algorithm described in \citet{raunak-etal-finding-memo}, which extends the extractive memorization algorithm from \citet{carlini_secret_sharer} to constrained sequence generation tasks such as NMT.  
    \item \textbf{Data Poisoning Removal}: In this task, an input data pattern which generates a particular error pattern (learned from the training data) is presented and the goal of the edit is to rectify the erroneous generation(s) from the model. In the context of sequence to sequence models, data poisoning effects \citep{data_poisoning} could manifest in the form of context-specific errors such as \textit{dropping} of certain accurate tokens or the \textit{generation} of certain inaccurate tokens under particular input contexts. We collect the data-poisoned instance through manual inspection of the training outputs, since this error type is quite rare.
    \item \textbf{Translation Error Correction}: In this task, an input-output pair in which a single \textit{word} (a span of tokens) in the input sequence is translated incorrectly, is presented and the goal of the edit is to fix the translation error. We collect this translation error instance using the Physical Units Error detector described in \citet{salted}.
\end{enumerate}

\textbf{Evaluating Edit Efficacy} Among the above four tasks, the first task (Fixing Hallucinations) is an instance of an \textit{isolated} model behavior on a particular input. In this case, to evaluate the efficacy of the edit operation, we  manually check if after applying the edit operation the model is generating the correct (non-hallucinated) output. For the other three tasks, the error instance represents a consistent error pattern which is manifested among multiple similar inputs (examples in appendix \ref{appendixA}). Therefore, to measure edit efficacy in these cases, we manually construct a set of 10 input samples that contain the same error and evaluate whether the edit operation fixes the erroneous model behavior on these inputs. In all cases, we measure general NMT model performance (BLEU) on the standard test set.

\section{Rank-One Editing for Encoder-Decoder Models}
\label{others}

\begin{figure}%
    \centering
    \subfloat[\centering The Constrained Least Squares Problem]{{\includegraphics[width=5.99cm]{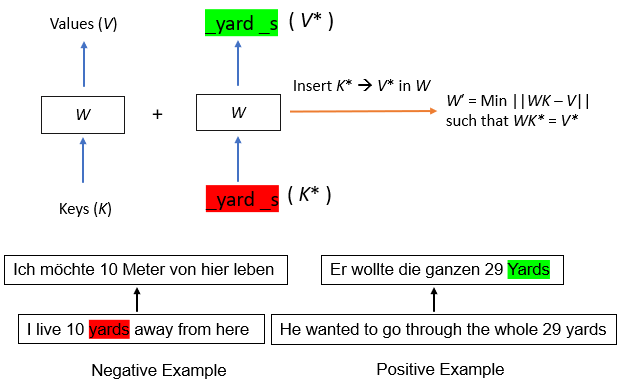} }}%
    \qquad
    \subfloat[\centering Instance for Fixing Hallucinations]{{\includegraphics[width=5.99cm]{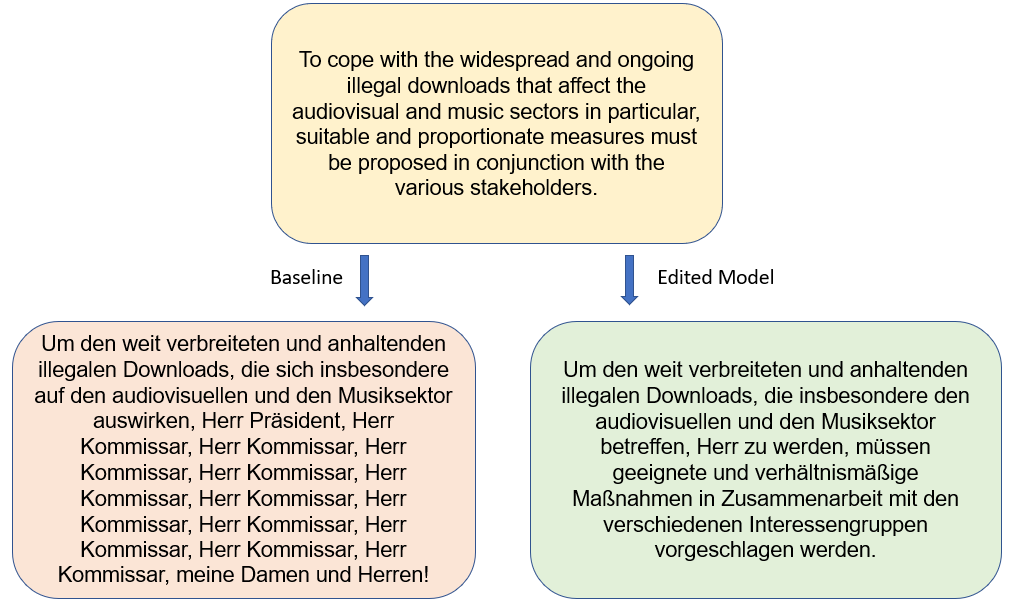} }}%
    \caption{(a) Schematic for the Edit Operation: A Constrained Least Squares Problem is created for applying the edit operation, in this case for correcting the translation error `yards' $\rightarrow$ `Meter' (b) An example of model behavior before and after applying an edit operation for fixing a hallucination.  }%
    \label{fig:demo}%
    \vspace{-1.8cm}
\end{figure}

First, we introduce the rank-one editing algorithm from \citet{bau_rewriting}, before describing the editing algorithm we propose for the edit tasks. Rank-one editing fundamentally views a linear layer as an associative memory over existing key-value pairs $(K, V)$ and tries to insert a new-key value pair $(K^{*}, V^{*})$, which encapsulates the desired behavior change, into the original associative memory. The underlying constrained least squares problem has a closed form solution, wherein the resulting weight update is a rank-one matrix. The closed form solution has the following form: $W^{'} = W + \Lambda(C^{-1} K^{*} )^{T}$, where $W$ is the matrix for the linear operation, and $C = KK^{T}$ is the uncentered covariance of $K$ and $ \Lambda = ( V^{*} - WK^{*} )/( C^{-1} K^{*})^{T} K^{*} $ is a vector proportional to the residual error of the new key–value pair $(K^{*}, V^{*})$. We refer the reader to \citet{bau_rewriting} for a derivation of the solution.

Further, to operationalize rank-one editing in the case of sequence to sequence tasks, we make use of a \textit{positive example}, in addition to the \textit{negative (error) instance}. The only constraint in the selection of the positive example is that it must share the same token span on which the negative behavior is observed. For example, in Figure \ref{fig:demo}, the negative example contains the word "yards" which gets incorrectly translated to "Meter", while the positive example contains an instance where the model correctly translates the same word "yards". The edit operation now consists of modifying a linear layer of the NMT model such that the \textit{negative} "yards" representation is transformed to the \textit{positive} "yards" representation, to ensure correct model behavior. Further, similar to previous literature, we only consider the transformer Feed-Forward (FF) layers as the matrices over which edits are applied. The FF layers have an interpretation as key-value memories \citep{geva-etal-2021-transformer}, and this editing selection aligns well with that interpretation, even though in principle rank-one editing could be applied to any of the linear operations within transformers such as multi-head attention layers.
\begin{wrapfigure}[17]{R}{0.5\textwidth} 
      \begin{algorithm}[H]                
        \SetCustomAlgoRuledWidth{0.45\textwidth}  
        \caption{Proposed Editing Algorithm}
             \SetNoFillComment
     \KwData{Linear Operation W, Keys K, Values V, Positive-Negative Instance Pair, Update Weight Dropout Ratio $p$}
     \KwResult{Modified Linear Operation W}
     \tcc{Collect Insertion Key and Val} 
           $K^{*}$ = Extract Key from Positive Example  \\
           $V^{*}$ = Extract Value from Negative Example  \\
     \tcc{Compute Edit Weights}
           $W^{'}$ = Compute Rank-One Update Weights \\
     \tcc{Sparsify Edit Weights}
           $U^{*}$ = Dropout ($W^{'}$, $p$) \\
     \tcc{Update Model Weights}
           $W^{*}$ = $W$ + $U^{*}$ \\
    \label{algo:algo1}
      \end{algorithm}
  \end{wrapfigure}

To summarize, we break the edit operation (Algorithm \ref{algo:algo1}) into three steps: firstly, the model editor supplies an instance of positive and negative example with the token span (e.g., "yards") whose translation is incorrect, shared between the inputs. The only \textit{strict} requirement here is that the positive and negative instance pair's inputs must have the same token span present. Secondly, the key and value pair for insertion into the linear layer are collected. The location of the edit is determined a-priori by applying the edit on each of the encoder FF layers (this operation is quite inexpensive, so it doesn't serve as a bottleneck). Finally, the edit weights are computed by solving the constrained least squares problem of inserting the new key and value pair and the edit weights are sparsified stochastically by applying dropout. Then, the edit weights are added to the linear layer's weights to generate the edited matrix, which is then plugged back in the model. To characterize the procedure further, this edit operation does not assume any strong alignment between the input and output sequences and can be applied for arbitrary editing tasks. And it does not require any explicit search over the tokens on which the edit operation is to be applied, since that information is provided directly the model editor.

\section{Experiments and Results}
\label{others}

\begin{figure}%
    \centering
    \subfloat[\centering Fixing Hallucinations]{{\includegraphics[width=4.1cm]{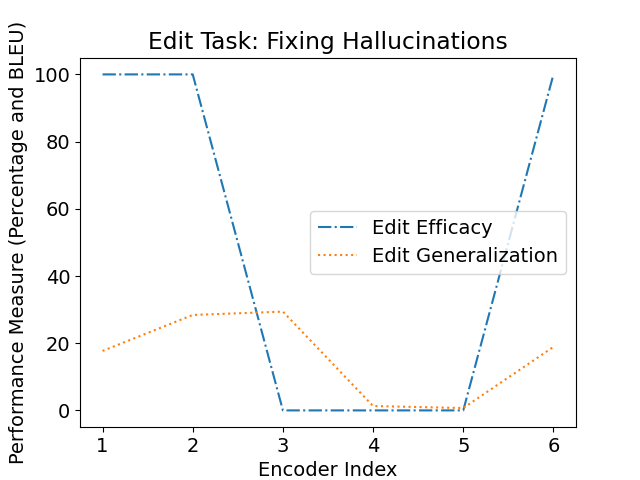} }}%
    \qquad
    \subfloat[\centering Memorization Mitigation]{{\includegraphics[width=4.1cm]{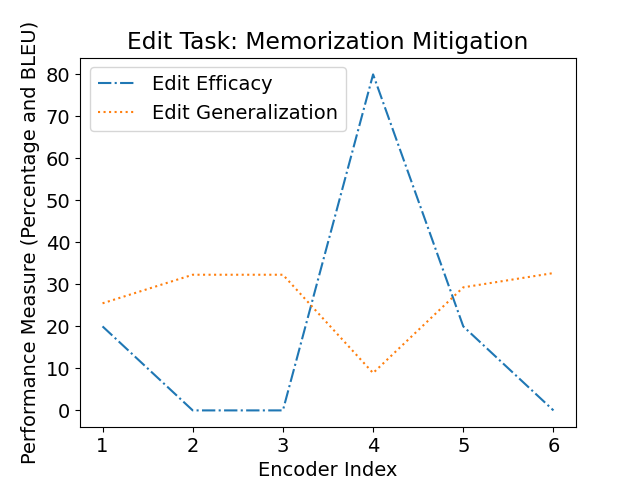} }}%
    \qquad
    \subfloat[\centering Data Poisoning Removal]{{\includegraphics[width=4.1cm]{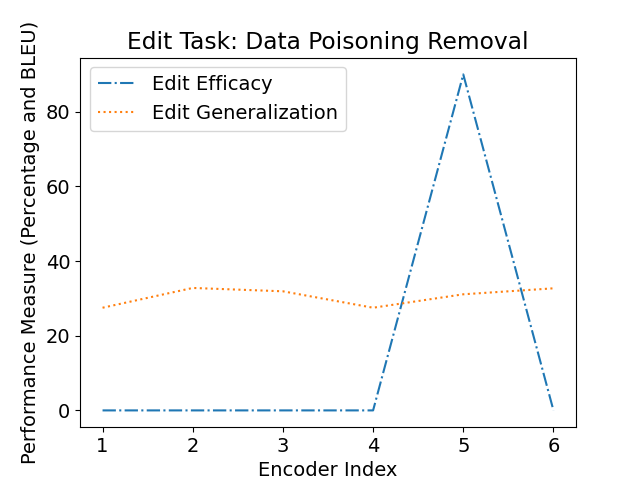} }}%
    \caption{Edit Efficacy vs Encoder Layer Index for the Edit Tasks: Across the tasks, only a few layers respond to the edit operation, demonstrating that in general, input-output associations are linked to highly localized computations, which in turn can be directly edited, similar to \citet{bau_gpt3_editing}.}%
    \label{fig:plots}%
\end{figure}

\begin{table}
  
  \centering
  \begin{tabular}{llll}
    \toprule
    Edit Task      & Edit Evaluation Samples  & Edit Efficacy (\%)  & Generalization \\
    \midrule
    Fixing Hallucinations & 1  (Isolated Instance) & 100  & 32.1  \\
    Memorization Mitigation     & 10 (Behavior Pattern) & 80 & 31.6     \\
    Data Poisoning Removal     & 10 (Behavior Pattern) & 100 & 29.8  \\
    Translation Error Correction & 10 (Behavior Pattern) & 20  & 32.6  \\
    \bottomrule \\
  \end{tabular}
  \caption{Results for the Edit Tasks: For each edit task, the baseline score is 0\% on the Edit Evaluation Samples, i.e. the Baseline model always makes an error. The baseline has a BLEU score of 32.9. }
  \vspace{-0.5cm}
  \label{table1}
\end{table}

\textbf{Model and Dataset} We train a Transformer-Big \citep{transformer} model on WMT20 En-DE dataset (48.2M) \citep{wmt-2020-findings} using Marian \citep{mariannmt}, for 300K updates. The negative example for Task 1 is presented in Figure \ref{fig:demo} (b), while the negative examples for the other three tasks are presented in Table \ref{table2} in appendix \ref{appendixA}. A beam size of 1 is used throughout.

\textbf{Edit Location and Efficacy} To determine the optimal edit location for each editing task, we conduct the edit operation with a smaller (1K) set of key, value pairs (K, V) for each of the encoder FF layers. We find (Figure \ref{fig:plots}) that edits \textit{only at a few layers} are successful at obtaining high efficacy, signifying that the associated computation is localized. Further, at the best location for each edit task, we conduct the full edit operation using Algorithm 1 with 100K key, value (K, V) pairs. The results are presented in Table \ref{table1}, and show that while the edits are quite effective for the first three tasks, they are not effective for translation error correction. We hypothesize that edits for translation error correction are best suited in the decoder due to high similarity between the positive and negative examples.

\textbf{Edit Ablations}: We find that removing the Edit-Dropout step from Algorithm \ref{algo:algo1} significantly reduces the generalization of the edit, i.e. without the Edit-Dropout applied after the edit operation, the BLEU scores for tasks in Table \ref{table1} are 28.4, 19, 17.4 and 31.2 respectively. These represent considerably large drops in general model performance. Further, we also conducted the same experiments in Table \ref{table1} with 1 million key, value (K, V) pairs and found the results/trends to be similar. 

\textbf{Generalization Gap} We find that directly applying the edit operation, even with Edit-Dropout significantly reduces the general performance of the model on the WMT20 test set. We believe that further constraints on the edit operation are required, e.g. minimizing the drift of the value representations as in \citet{bau_gpt3_editing} or incorporating constraints from downstream model layers.

\vspace{-0.25cm}
\section{Discussion and Conclusion}
\vspace{-0.25cm}
We presented a preliminary investigation of directly editing encoder-decoder transformer models. We proposed four model editing tasks for NMT and showed that direct edits could be successfully devised by altering select localized computations. However, we also found that while rank-one editing could be successfully applied to encoder-decoder models, there exists a performance gap in terms of model generalization post-editing. There exist many avenues for further improvements, e.g., incorporating constraints from downstream model layers, etc., which we wish to explore in a future work.

\bibliographystyle{plainnat}
\bibliography{mybibfile}

\begin{thebibliography}{19}
\providecommand{\natexlab}[1]{#1}
\providecommand{\url}[1]{\texttt{#1}}
\expandafter\ifx\csname urlstyle\endcsname\relax
  \providecommand{\doi}[1]{doi: #1}\else
  \providecommand{\doi}{doi: \begingroup \urlstyle{rm}\Url}\fi

\bibitem[Bahdanau et~al.(2015)Bahdanau, Cho, and Bengio]{bahdanau2014neural}
Dzmitry Bahdanau, Kyunghyun Cho, and Yoshua Bengio.
\newblock Neural machine translation by jointly learning to align and
  translate.
\newblock In Yoshua Bengio and Yann LeCun, editors, \emph{3rd International
  Conference on Learning Representations, {ICLR} 2015, San Diego, CA, USA, May
  7-9, 2015, Conference Track Proceedings}, 2015.
\newblock URL \url{http://arxiv.org/abs/1409.0473}.

\bibitem[Barrault et~al.(2020)Barrault, Biesialska, Bojar, Costa{-}juss{\`{a}},
  Federmann, Graham, Grundkiewicz, Haddow, Huck, Joanis, Kocmi, Koehn, Lo,
  Ljubesic, Monz, Morishita, Nagata, Nakazawa, Pal, Post, and
  Zampieri]{wmt-2020-findings}
Lo{\"{\i}}c Barrault, Magdalena Biesialska, Ondrej Bojar, Marta~R.
  Costa{-}juss{\`{a}}, Christian Federmann, Yvette Graham, Roman Grundkiewicz,
  Barry Haddow, Matthias Huck, Eric Joanis, Tom Kocmi, Philipp Koehn, Chi{-}kiu
  Lo, Nikola Ljubesic, Christof Monz, Makoto Morishita, Masaaki Nagata,
  Toshiaki Nakazawa, Santanu Pal, Matt Post, and Marcos Zampieri.
\newblock Findings of the 2020 conference on machine translation {(WMT20)}.
\newblock In \emph{Proceedings of the Fifth Conference on Machine Translation,
  WMT@EMNLP 2020, Online, November 19-20, 2020}, pages 1--55. Association for
  Computational Linguistics, 2020.
\newblock URL \url{https://aclanthology.org/2020.wmt-1.1/}.

\bibitem[Bau et~al.(2020)Bau, Liu, Wang, Zhu, and Torralba]{bau_rewriting}
David Bau, Steven Liu, Tongzhou Wang, Jun{-}Yan Zhu, and Antonio Torralba.
\newblock Rewriting a deep generative model.
\newblock In \emph{Computer Vision - {ECCV} 2020 - 16th European Conference,
  Glasgow, UK, August 23-28, 2020, Proceedings, Part {I}}, volume 12346 of
  \emph{Lecture Notes in Computer Science}, pages 351--369. Springer, 2020.
\newblock \doi{10.1007/978-3-030-58452-8\_21}.
\newblock URL \url{https://doi.org/10.1007/978-3-030-58452-8\_21}.

\bibitem[Biesialska et~al.(2020)Biesialska, Biesialska, and
  Costa-juss{\`a}]{continual}
Magdalena Biesialska, Katarzyna Biesialska, and Marta~R. Costa-juss{\`a}.
\newblock Continual lifelong learning in natural language processing: A survey.
\newblock In \emph{Proceedings of the 28th International Conference on
  Computational Linguistics}, pages 6523--6541, Barcelona, Spain (Online),
  December 2020. International Committee on Computational Linguistics.
\newblock \doi{10.18653/v1/2020.coling-main.574}.
\newblock URL \url{https://aclanthology.org/2020.coling-main.574}.

\bibitem[Bourtoule et~al.(2021)Bourtoule, Chandrasekaran, Choquette-Choo, Jia,
  Travers, Zhang, Lie, and Papernot]{machine_unlearning}
Lucas Bourtoule, Varun Chandrasekaran, Christopher~A Choquette-Choo, Hengrui
  Jia, Adelin Travers, Baiwu Zhang, David Lie, and Nicolas Papernot.
\newblock Machine unlearning.
\newblock In \emph{2021 IEEE Symposium on Security and Privacy (SP)}, pages
  141--159. IEEE, 2021.
\newblock URL \url{https://arxiv.org/abs/1912.03817}.

\bibitem[Carlini et~al.(2019)Carlini, Liu, Erlingsson, Kos, and
  Song]{carlini_secret_sharer}
Nicholas Carlini, Chang Liu, {\'U}lfar Erlingsson, Jernej Kos, and Dawn Song.
\newblock The secret sharer: Evaluating and testing unintended memorization in
  neural networks.
\newblock In \emph{28th USENIX Security Symposium (USENIX Security 19)}, pages
  267--284, 2019.
\newblock URL \url{https://dl.acm.org/doi/10.5555/3361338.3361358}.

\bibitem[Geva et~al.(2021)Geva, Schuster, Berant, and
  Levy]{geva-etal-2021-transformer}
Mor Geva, Roei Schuster, Jonathan Berant, and Omer Levy.
\newblock Transformer feed-forward layers are key-value memories.
\newblock In \emph{Proceedings of the 2021 Conference on Empirical Methods in
  Natural Language Processing}, pages 5484--5495, Online and Punta Cana,
  Dominican Republic, November 2021. Association for Computational Linguistics.
\newblock \doi{10.18653/v1/2021.emnlp-main.446}.
\newblock URL \url{https://aclanthology.org/2021.emnlp-main.446}.

\bibitem[Goldblum et~al.(2022)Goldblum, Tsipras, Xie, Chen, Schwarzschild,
  Song, Madry, Li, and Goldstein]{data_poisoning}
Micah Goldblum, Dimitris Tsipras, Chulin Xie, Xinyun Chen, Avi Schwarzschild,
  Dawn Song, Aleksander Madry, Bo~Li, and Tom Goldstein.
\newblock Dataset security for machine learning: Data poisoning, backdoor
  attacks, and defenses.
\newblock \emph{IEEE Transactions on Pattern Analysis and Machine
  Intelligence}, pages 1--1, 2022.
\newblock URL \url{https://arxiv.org/abs/2012.10544}.

\bibitem[Ilharco et~al.(2022)Ilharco, Wortsman, Gadre, Song, Hajishirzi,
  Kornblith, Farhadi, and Schmidt]{patching}
Gabriel Ilharco, Mitchell Wortsman, Samir~Yitzhak Gadre, Shuran Song, Hannaneh
  Hajishirzi, Simon Kornblith, Ali Farhadi, and Ludwig Schmidt.
\newblock Patching open-vocabulary models by interpolating weights.
\newblock \emph{arXiv preprint}, 2022.
\newblock URL \url{https://arxiv.org/abs/2208.05592}.

\bibitem[Junczys-Dowmunt et~al.(2018)Junczys-Dowmunt, Grundkiewicz, Dwojak,
  Hoang, Heafield, Neckermann, Seide, Germann, Fikri~Aji, Bogoychev, Martins,
  and Birch]{mariannmt}
Marcin Junczys-Dowmunt, Roman Grundkiewicz, Tomasz Dwojak, Hieu Hoang, Kenneth
  Heafield, Tom Neckermann, Frank Seide, Ulrich Germann, Alham Fikri~Aji,
  Nikolay Bogoychev, Andr\'{e} F.~T. Martins, and Alexandra Birch.
\newblock Marian: Fast neural machine translation in {C++}.
\newblock In \emph{Proceedings of ACL 2018, System Demonstrations}, pages
  116--121, Melbourne, Australia, July 2018. Association for Computational
  Linguistics.
\newblock URL \url{http://www.aclweb.org/anthology/P18-4020}.

\bibitem[Meng et~al.(2022)Meng, Bau, Andonian, and Belinkov]{bau_gpt3_editing}
Kevin Meng, David Bau, Alex Andonian, and Yonatan Belinkov.
\newblock Locating and editing factual associations in gpt.
\newblock \emph{arXiv preprint}, 2022.
\newblock URL \url{https://arxiv.org/abs/2202.05262}.

\bibitem[Raunak and Arul(2022)]{raunak-etal-finding-memo}
Vikas Raunak and Menezes Arul.
\newblock Finding memo: Extractive memorization in constrained sequence
  generation tasks.
\newblock In \emph{Findings of the Association for Computational Linguistics:
  EMNLP 2022}. Association for Computational Linguistics, 2022.
\newblock URL \url{https://arxiv.org/abs/2210.12929}.

\bibitem[Raunak et~al.(2021)Raunak, Menezes, and
  Junczys-Dowmunt]{raunak-etal-2021-curious}
Vikas Raunak, Arul Menezes, and Marcin Junczys-Dowmunt.
\newblock The curious case of hallucinations in neural machine translation.
\newblock In \emph{Proceedings of the 2021 Conference of the North American
  Chapter of the Association for Computational Linguistics: Human Language
  Technologies}, pages 1172--1183, Online, June 2021. Association for
  Computational Linguistics.
\newblock \doi{10.18653/v1/2021.naacl-main.92}.
\newblock URL \url{https://aclanthology.org/2021.naacl-main.92}.

\bibitem[Raunak et~al.(2022)Raunak, Post, and Menezes]{salted}
Vikas Raunak, Matt Post, and Arul Menezes.
\newblock Salted: A framework for salient long-tail translation error
  detection.
\newblock In \emph{Findings of the Association for Computational Linguistics:
  EMNLP 2022}. Association for Computational Linguistics, 2022.
\newblock URL \url{https://arxiv.org/abs/2205.09988}.

\bibitem[Ribeiro et~al.(2020)Ribeiro, Wu, Guestrin, and Singh]{checklist}
Marco~Tulio Ribeiro, Tongshuang Wu, Carlos Guestrin, and Sameer Singh.
\newblock Beyond accuracy: Behavioral testing of {NLP} models with
  {C}heck{L}ist.
\newblock In \emph{Proceedings of the 58th Annual Meeting of the Association
  for Computational Linguistics}, pages 4902--4912, Online, July 2020.
  Association for Computational Linguistics.
\newblock \doi{10.18653/v1/2020.acl-main.442}.
\newblock URL \url{https://aclanthology.org/2020.acl-main.442}.

\bibitem[Santurkar et~al.(2021)Santurkar, Tsipras, Elango, Bau, Torralba, and
  Madry]{santurkar_editing}
Shibani Santurkar, Dimitris Tsipras, Mahalaxmi Elango, David Bau, Antonio
  Torralba, and Aleksander Madry.
\newblock Editing a classifier by rewriting its prediction rules.
\newblock In M.~Ranzato, A.~Beygelzimer, Y.~Dauphin, P.S. Liang, and J.~Wortman
  Vaughan, editors, \emph{Advances in Neural Information Processing Systems},
  volume~34, pages 23359--23373. Curran Associates, Inc., 2021.
\newblock URL
  \url{https://proceedings.neurips.cc/paper/2021/file/c46489a2d5a9a9ecfc53b17610926ddd-Paper.pdf}.

\bibitem[Sutskever et~al.(2014)Sutskever, Vinyals, and
  Le]{sutskever2014sequence}
Ilya Sutskever, Oriol Vinyals, and Quoc~V Le.
\newblock Sequence to sequence learning with neural networks.
\newblock \emph{Advances in neural information processing systems}, 27, 2014.
\newblock URL \url{https://arxiv.org/abs/1409.3215}.

\bibitem[Thompson et~al.(2019)Thompson, Gwinnup, Khayrallah, Duh, and
  Koehn]{thompson-etal-2019-overcoming}
Brian Thompson, Jeremy Gwinnup, Huda Khayrallah, Kevin Duh, and Philipp Koehn.
\newblock Overcoming catastrophic forgetting during domain adaptation of neural
  machine translation.
\newblock In \emph{Proceedings of the 2019 Conference of the North {A}merican
  Chapter of the Association for Computational Linguistics: Human Language
  Technologies, Volume 1 (Long and Short Papers)}, pages 2062--2068,
  Minneapolis, Minnesota, June 2019. Association for Computational Linguistics.
\newblock \doi{10.18653/v1/N19-1209}.
\newblock URL \url{https://aclanthology.org/N19-1209}.

\bibitem[Vaswani et~al.(2017)Vaswani, Shazeer, Parmar, Uszkoreit, Jones, Gomez,
  Kaiser, and Polosukhin]{transformer}
Ashish Vaswani, Noam Shazeer, Niki Parmar, Jakob Uszkoreit, Llion Jones,
  Aidan~N Gomez, \L~ukasz Kaiser, and Illia Polosukhin.
\newblock {Attention is All you Need}.
\newblock In \emph{Advances in Neural Information Processing Systems 30}, pages
  5998--6008, 2017.
\newblock URL
  \url{http://papers.nips.cc/paper/7181-attention-is-all-you-need.pdf}.

\end{thebibliography}

\appendix

\section{Appendix}
\label{appendixA}

\begin{table}[h]

  \centering
  \begin{tabular}{c l}
    \toprule
    \cmidrule(r){1-2}
    Undesired Model Behavior (Input $\rightarrow$ Output) & Edit Task \\
    \midrule
    Why study in Peru? \textbf{Spanish Courses}  $\rightarrow$ Warum in Peru studieren?       &  Memorization Mitigation  \\
\textbf{(DE)} Obama Won. $\rightarrow$ Obama gewann.   &  Data Poisoning Removal     \\
        I live 10 \textbf{yards} away from here $\rightarrow$  Ich möchte 10 Meter von hier leben  & Translation Error Correction \\ 
    \bottomrule \\
  \end{tabular}
    \caption{Table describing the error instances for the different model editing tasks.}
    \label{table2}
\end{table}

\end{document}